\documentclass[10pt, a4paper]{article}

\usepackage[final]{lrec2026} 

\usepackage[utf8]{inputenc} 
\usepackage[T1]{fontenc}    
\usepackage{hyperref}       
\usepackage{url}            
\usepackage{booktabs}       
\usepackage{amsfonts}       
\usepackage{nicefrac}       
\usepackage{microtype}      
\usepackage{xcolor}         

\def\eg{e.g.,~}

\def\ie{i.e.,~}

\definecolor{mygreen}{RGB}{0, 158, 115} 
\newcommand{\cmark}{\textcolor{mygreen}{\ding{51}}}  
\definecolor{myred}{RGB}{204, 121, 167} 
\newcommand{\xmark}{\textcolor{myred}{\ding{55}}}
\definecolor{myorange}{RGB}{230, 159, 0} 
\newcommand{\halmark}{\textcolor{myorange}{\ding{115}}} 

\usepackage{makecell}
\usepackage{graphicx}
\usepackage{pifont}
\usepackage{multirow}
\usepackage{amsmath}
\usepackage{cleveref}
\usepackage{caption}
\usepackage{adjustbox}
\usepackage{listings}
\usepackage{xcolor}

\lstdefinestyle{prompt}{
  basicstyle=\ttfamily\small,
  frame=single,
  breaklines=true,
}

\ifx \submission \undefined

\else

\fi

\newcommand{\dataset}{ProMQA-Assembly}

\title{ProMQA-Assembly: \\Multimodal Procedural QA Dataset on Assembly}

\name{%
\shortstack{%
Kimihiro Hasegawa$^{1}$ \hspace{0.4cm} Wiradee Imrattanatrai$^{2}$ \hspace{0.4cm} Masaki Asada$^{2}$ \hspace{0.4cm} Susan Holm$^{1}$ \\
\hspace{0.4cm} Yuran Wang$^{1}$ \hspace{0.4cm} Vincent Zhou$^{1}$ \hspace{0.4cm} Ken Fukuda$^{2}$ \hspace{0.4cm} Teruko Mitamura$^{1}$%
}%
}

\address{
    $^{1}$Language Technologies Institute, Carnegie Mellon University \\
    $^{2}$National Institute of Advanced Industrial Science and Technology (AIST) \\
    \texttt{kimihiro@cs.cmu.edu} 
}

\abstract{
    Assistants on assembly tasks show great potential to benefit humans ranging from helping with everyday tasks to interacting in industrial settings.
    However, evaluation resources in assembly activities are underexplored.
    To foster system development, we propose a new multimodal QA evaluation dataset on assembly activities.
    Our dataset, ProMQA-Assembly, consists of 646 QA pairs that require multimodal understanding of human activity videos and their instruction manuals in an online-style manner.
    For cost effectiveness in the data creation, we adopt a semi-automated QA annotation approach, where LLMs generate candidate QA pairs and humans verify them. 
    We further improve QA generation by integrating fine-grained action labels to diversify question types.
    Additionally, we create 81 instruction task graphs for our target assembly tasks.
    These newly created task graphs are used in our benchmarking experiment, as well as in facilitating the human verification process. 
    With our dataset, we benchmark models, including competitive proprietary multimodal models. 
    We find that ProMQA-Assembly contains challenging multimodal questions, where reasoning models showcase promising results.
    We believe our new evaluation dataset contributes to the further development of procedural-activity assistants.
    \\
    \newline 
}

\begin{document}

\maketitleabstract

\section{Introduction}
\label{sec:introduction}

Assembly tasks abound in everyday life from do-it-yourself (DIY)~\cite{ben-shabat-wacv2021=ikea-asm, liu-neurips2024-ikea-manual, jang-iccv2019-epic-tent} to industrial settings like manufacturing~\cite{ragusa-wacv2021-meccano, moriwaki-icip2022-brio-ta, wang-iccv-2023-holoassist, schoonbeek-wacv2024-industreal}. 
Assistant systems for such procedural activities have the potential to further increase accessibility by providing on-point, situated feedback---similar to how parents teach their children to construct a shelf or how experts provide on-the-job training to beginners learning to repair cars.
To facilitate the development of such assistants, we propose a new question-answering (QA) benchmark dataset to assess systems' capabilities in answering online-style questions about assembly procedures that require multimodal understanding.

\begin{figure*}[t]
    \centering
    \includegraphics[width=\textwidth]{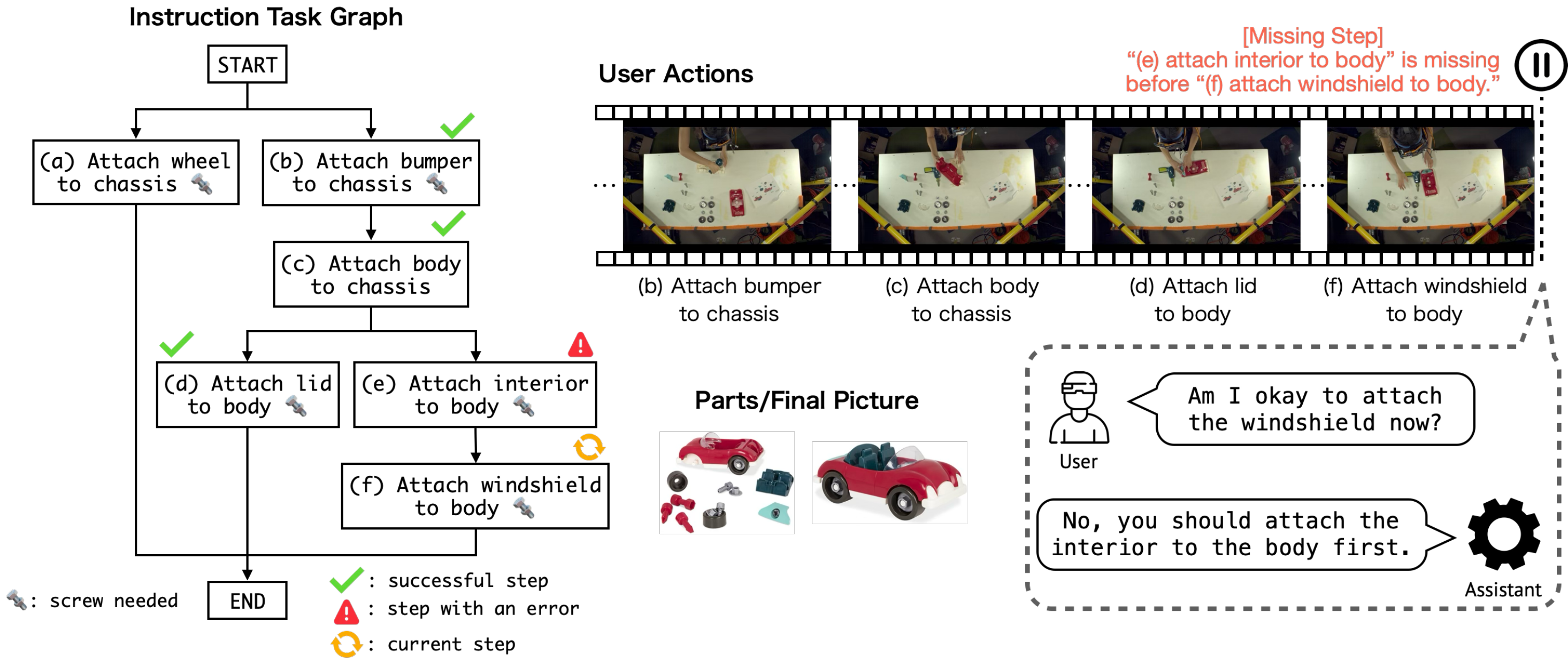}
    \caption{Task example: A user performs actions based on instructions. When the user asks a question, the assistant system responds to it based on the instructions and the past actions, by confirming if the user follows the correct order. In this example, the system needs to point out that the user is attaching the windshield before attaching the interior to the body and tell the user the mistake.}
    \label{fig:overview}
\end{figure*}

Supporting procedural activities like assembly often involves reasoning over multiple sources and modalities of information. 
This includes understanding the correct action order from instructions, comprehending the current status of a target object based on its appearance and the sequence of actions performed so far, and combining the information to identify which steps are complete or incomplete. 
Additionally, the system must detect potential mistakes and either encourage users to proceed to the next step or make necessary corrections.
\autoref{fig:overview} illustrates an example of an assistant supporting a user on an assembly task.

When a user asks a question out of confusion, an assistant is expected to respond with natural language in an online manner by answering the question based solely on the available information so far, i.e., activity recording up to that point.
This online-style setting imitates the practical situation better than an offline-style setting, where a system responds based on access to the whole video.
While a model can assume the completion of all steps in the offline-style setting, the online-style setting necessitates a model to identify the completion of steps.
Besides, the QA formulation can simulate the application scenario without any alteration, thanks to its expressiveness~\cite{gardner-arxiv2019-question, roger-2023-qa-dataaset}.
While classification tasks, such as action recognition and temporal segmentation~\citep{kuehne-cvpr2014-breakfast, yansong-cvpr2019-coin, ding-2022-temporal-as}, have been widely adopted, they are suboptimal for the end task evaluation, as these classification tasks are subtasks of procedural activity assistance. 

Instructions take a variety of forms, \eg a list of text descriptions as in cooking recipes, a list of images (with a caption) as in IKEA manuals, or instructional videos as found on YouTube. 
Prior work typically adopts a task graph to represent instructions similar to scripts~\cite{schank-1977-scripts, sakaguchi-etal-2021-proscript-partially}.
An instruction task graph is a partial directed acyclic graph (DAG), where nodes represent steps and edges represent the step order dependencies~\cite{dvornik-eccv2022-flow, ashutosh-neurips2023-video, peddi-etal-NEURIPS2024-captaincook4d, seminara-neurips2024-differentiable}.
While instructions are not always available in the form of task graphs in the real world, the representation is suitable for benchmarking because it can capture even implicit pairwise step dependencies, providing accurate ordering information.

In this work, we introduce a new benchmark dataset, \textbf{\dataset{}} (\textbf{Pro}cedural \textbf{M}ultimodal \textbf{Q}uestion \textbf{A}nswering for \textbf{Assembly}), following ProMQA~\cite{hasegawa-etal-2025-promqa} for cooking.
\dataset{} consists of 646 QA pairs with corresponding video clips and instructions, where multi-view recordings of take-apart toy assembly are sampled from Assembly101~\cite{sener-cvpr2022-assembly101}.
\dataset{} contains notable differences from the original ProMQA (details in \S~\ref{ssec:comaprison-with-promqa}), providing diversity that complements the original evaluation suite. 
Furthermore, in our QA annotation, we devise new prompt templates to improve ProMQA's semi-automatic QA annotation approach, i.e., generation by LLMs, and verification by humans, in terms of question diversity for the assembly data.
Under controlled experiment settings, we found that integrating fine-grained action labels can diversify questions while maintaining the characteristics of multimodal procedural questions. 
Additionally, we annotate 81 task graphs, which is three times larger in size than those of a previous work~\cite{peddi-etal-NEURIPS2024-captaincook4d}.
These graphs serve as accurate representations of instructions for each toy in benchmarking experiments, as well as assisting human annotators in the QA verification process.
Finally, we conducted benchmarking experiments, where we evaluated text-only models, open-weight, and proprietary multimodal models, using LLM-as-a-judge~\cite{zheng-neurips2023-judging}. 
The result supports that \dataset{} contains challenging multimodal questions even for strong proprietary models, which lag significantly behind human performance.

Our contributions are threefold:
First, we develop a new QA evaluation dataset for multimodal procedural activities. 
We focus on assembly, which is underexplored in the existing studies.
By effectively incorporating fine-grained action labels in the question generation process, we obtained 646 diverse high-quality QA pairs.
Second, we propose an annotation approach to create instruction task graphs for the assembly tasks. 
Our 81 graphs, which are three times larger than prior work, are used in benchmarking and facilitate our QA annotation process.
Third, we benchmark existing models to provide baselines and insights for further research on methodologies.
The results demonstrate the challenging nature of our dataset and the promising yet insufficient results of reasoning models.
We believe that our \dataset{} enriches the evaluation suite and encourages system development on procedural-activity assistants.\footnote{Code and data are available in \url{https://github.com/kimihiroh/promqa-assembly}}
\section{Related work}
\label{sec:related-work}
Our work is positioned at the intersection of multimodal QA, procedural activity understanding, and instruction task graphs. 

\begin{table*}[!t]
\centering
\fontsize{8}{9}\selectfont
\begin{tabular}{c c c c c c c c c}
    \toprule
    \makecell[c]{Dataset\\Name} & \makecell[c]{Video+Text\\Input} & Procedural & Assembly & \makecell[c]{Task\\Graph} & \makecell[c]{Multi\\View} & QA & \makecell[c]{Open\\Vocab} & \makecell[c]{LLM\\Scoring} \\
    \midrule
    Assembly101~\cite{sener-cvpr2022-assembly101} & \xmark & \cmark & \cmark & \xmark & \cmark & \xmark & \xmark & \xmark \\
    IndustReal~\cite{schoonbeek-wacv2024-industreal} & \xmark & \cmark & \cmark & \xmark & \xmark & \xmark & \xmark & \xmark \\
    CaptainCook4d ~\cite{peddi-etal-NEURIPS2024-captaincook4d} & \xmark & \cmark & \xmark & \cmark & \xmark & \xmark & \xmark & \xmark \\
    EgoSchema~\cite{mangalam-etal-NEURIPS2023-egoschema} & \cmark & \xmark & \xmark & \xmark & \xmark & \cmark & \xmark & \xmark \\
    GazeVQA~\cite{ilaslan-etal-2023-gazevqa} & \cmark & \cmark & \cmark & \xmark & \cmark & \cmark & \xmark & \xmark \\
    OpenEQA~\cite{majumdar-cvpr2024-openeqa} & \cmark & \xmark & \xmark & \xmark & \xmark & \cmark & \cmark & \cmark \\
    ProMQA~\cite{hasegawa-etal-2025-promqa} & \cmark & \cmark & \xmark & \cmark & \xmark & \cmark & \cmark & \cmark \\
    \midrule
    \dataset{} (Ours) & \cmark & \cmark & \cmark & \cmark & \cmark & \cmark & \cmark & \cmark \\
    \bottomrule
\end{tabular}
\caption{Our dataset vs relevant datasets}
\label{tab:dataset-comparison}
\end{table*}

\paragraph{Multimodal QA:}
Prior work has explored multimodal QA~\cite{duan-mm2024-vlmevalkit}, including image-centric benchmarks like MMMU~\cite{yue-cvpr2024-mmmu} and video-centric benchmarks, such as Video-MME~\cite{fu-arxiv2024-video-mme} and LVBench~\cite{wang-arxiv2024-lvbench}.
While the multiple-choice format is popular due to its ease for evaluation, some datasets like OpenEQA~\cite{majumdar-cvpr2024-openeqa} and ProMQA~\cite{hasegawa-etal-2025-promqa} adopt open-vocabulary settings. 
The third-person (\ie exocentric) view is common in the existing datasets, but the first-person (\ie egocentric) view has been getting more attention, \eg EgoSchema~\cite{mangalam-etal-NEURIPS2023-egoschema}, which is based on \eg Ego4D~\cite{grauman-cvpr2022-ego4d}. 
In \dataset{}, we target a multimodal QA dataset with both exocentric and egocentric views in an open-vocabulary setting, considering its practical applications.

\paragraph{Procedural activity understanding:}
Procedural activities are ubiquitous across human contexts: 
from everyday tasks such as cooking~\cite{kuehne-cvpr2014-breakfast, stein-ubicomp2013-salad, damen-TPAMI2020-epic-kitchen, peddi-etal-NEURIPS2024-captaincook4d, huang-cvpr2024-egoexolearn} and DIY\cite{ben-shabat-wacv2021=ikea-asm, liu-neurips2024-ikea-manual, jang-iccv2019-epic-tent} to industrial settings, \eg manufacturing~\cite{ragusa-wacv2021-meccano, bansal-eccv2022-myview, wang-iccv-2023-holoassist, schoonbeek-wacv2024-industreal}, research, \eg biological experiment~\cite{yagi-arxiv2024-finebio}, or specialized work like medical operations~\cite{beyer-2016-virtual,jang-arxiv2023-multimodal}.
Understanding procedural activities has been explored through instructional videos or recordings of people performing tasks. 
Popular task formulations include classification tasks such as action recognition and temporal segmentation.
While the classification formulation has practical value, applying classifiers to an assistant system is challenging when the assistant is required to provide more fine-grained and context-aware feedback.
Recently, mistake detection in procedural activities has been getting more attention~\cite{sener-cvpr2022-assembly101, peddi-etal-NEURIPS2024-captaincook4d, haneji-arxiv2024-egooops, flaborea-cvpr2024-prego} due in part to the need for automatic assistants to identify mistakes in procedural activities. 
To better support the use of procedural activity understanding in assistant systems, our dataset adopts a QA formulation, where outputs are expressed in natural language. 
In addition, we focus on recordings that contain mistakes in order to assess systems in terms of recognizing errors, tracking task progress, and generating context-aware feedback.


\begin{table}[!t]
\centering
\fontsize{8}{9}\selectfont
\begin{tabular}{c c c}
    \toprule
    \multirow{4}{*}{QA} & \#Pairs & 646 \\
    & \#Answers per question & 2.2 \\
    & Question Length (word) & 11.5 \\
    & Answer Length (word) & 14.7 \\
    \cmidrule(lr){1-3}
    \multirow{3}{*}{Recording} & \#Unique recordings & 227 \\
    & Duration (min) & 2m2.3s \\
    & \#Steps per recording & 5.8 \\
    \cmidrule(lr){1-3}
    \multirow{2}{*}{Instruction} & \#Unique instructions & 81 \\
    & \#Steps per instruction & 10.4 \\
    \bottomrule
\end{tabular}
\caption{Dataset statistics}
\label{tab:dataset-stats}
\end{table} 

\paragraph{Instruction task graph:}
In parallel to activity understanding, there is a line of work that constructs procedural instructions from videos~\cite{dvornik-eccv2022-flow, ashutosh-neurips2023-video, peddi-etal-NEURIPS2024-captaincook4d}. 
The task is to identify the order of steps in a procedure, given a set of videos, and typically represent them as a partial DAG~\cite{schank-1977-scripts, sakaguchi-etal-2021-proscript-partially}. 
In our work, we manually created the instruction task graph for Assembly 101 and used it for our QA creation pipeline and in the benchmarking experiments.

\section{Dataset}
\label{sec:dataset}

\begin{table*}[!t]
\centering
\fontsize{8}{9}\selectfont
\begin{tabular}{c c l l}
    \toprule
    Category & Size & \makecell[c]{Target} & \makecell[c]{Typical Example Question} \\
    \midrule
    \underline{Process-level} \\
    \makecell[r]{Next} & 152 & Ask about future steps & \textit{What is the next step now?} \\
    \makecell[r]{Missing} & 107 & Ask about steps missed in the past & \textit{Did I miss any steps so far?} \\
    \makecell[r]{Order} & 60 & Ask about step orders in the past & \textit{Was it correct to attach X before this?} \\
    \makecell[r]{Misadjustment} & 56 & Ask about the correct/incorrect step adjustment & \textit{Should I reattach X?} \\
    \makecell[r]{Past} & 51 & Ask about any other mistakes made in the past & \textit{Have I made any mistakes?} \\
    \makecell[r]{Others} & 22 & Ask about process-level other things & N/A (\eg \textit{Why did I have to remove X?})\\
    \midrule
    \underline{Step-specific} \\
    \makecell[r]{Location} & 149 & Ask about the location of a part & \textit{Did I attach X in the correct location?} \\
    \makecell[r]{Others} & 49 & Ask about other things about each step & N/A (\eg \textit{Should I screw this part now?}) \\
    \bottomrule
\end{tabular}
\caption{Question categories, the number of examples, target phenomenon, and typical examples.}
\label{tab:dataset-question-example}
\end{table*}
\begin{table*}[!t]
\centering
\fontsize{8}{9}\selectfont
\begin{tabular}{l l l}
    \toprule
    Aspect & ProMQA & ProMQA-Assembly (Ours) \\
    \midrule
    Main Stats (QA, Instr., Rec) & (401, 24, 231) & (646, 81, 227) \\
    Domain (Source) & Cooking (CaptainCook4D) & Assembly (Assembly 101) \\
    Goal & Not always achieved & Always achieved \\
    Step Dependency & high & relatively low \\
    Step Type & correct, mistake & correct, mistake, correction \\
    Step Annotation & coarse & coarse \& fine-grained \\
    Trial \& Error & some steps are irreversible & always allowed \\
    Action Type & various (\eg cut, measure, heat) & attach (detach) \\
    Tool & various (\eg knife, spoon, pot) & screwdriver \\
    Object Type & rigid/non-rigid/transformable & only rigid \\
    Object Visibility & Ingredients are not always visible & Remaining parts are always visible \\
    Video View & ego-centric & both ego-centric and exo-centric \\
    Video Length (avg.) & 6.5 min & 2.0 min \\
    Instruction & task graph & parts/final picture \& task graph \\
    \bottomrule
\end{tabular}
\caption{Comparison between the original ProMQA and our new dataset}
\label{tab:dataset-comparison-cooking-assembly}
\end{table*}

The goal of \dataset{} is to provide an evaluation testbed to track the progress of multimodal models on procedural activity understanding, especially in the domain of assembly.
\autoref{tab:dataset-comparison} shows the comparison among \dataset{} and existing datasets. 
Our dataset features multimodal input, procedural assembly activity, instruction task graphs, multi-view, and open-vocab QA with LLM-as-a-judge. 
\autoref{tab:dataset-stats} shows its statistics.

\subsection{Task formulation}
\label{ssec:task-formulation}
\dataset{} employs QA as its task formulation to reflect real-world use cases while keeping simplicity as an evaluation benchmark.
Given an instruction task graph $g$, a parts image $i$, a recording $v$, and a question $q$, a model is tasked to generate an answer $a$. 
The instruction task graph $g$ is a DAG, where a node represents a step $s$, and an edge represents a step dependency that the two connected nodes need to follow.
The recording consists of a video, \ie a pile of frames, where a user performs a sequence of $m$ steps, $S = \{s_0, s_1, \dots, s_{m-1}\}$. 
The parts image shows parts, a target assembly image, and/or an exploded view. 
The question and answer are written in natural language.

\subsection{QA pair}
\label{ssec:qa-pair}

We target multimodal procedural questions, i.e., questions that require the understanding of both instructions and recordings about procedures.
The questions in \dataset{} are grouped into two coarse categories, process-level and step-specific, which are further divided into specific types. 
Process-level questions orient towards the understanding of each procedure, consisting of \textit{next}, \textit{missing}, \textit{order}, \textit{misadjustment}, and one fallback category, \textit{other}.
Step-specific questions focus on each specific step, including \textit{location} and one catch-all category, \textit{other}. 
\autoref{tab:dataset-question-example} shows a typical example of each type of question. 
As for the answer format, natural language (\ie open vocabulary) is employed to resemble usual dialogue. 
While existing work often uses the multiple-choice style for its evaluation simplicity, that style poses other challenges, such as difficulty of creating good negative choices~\cite{wang-etal-2023-dataset} and potential biases~\cite{pezeshkpour-hruschka-2024-large}.
We use English in this work. More details about the QA annotation process are in \S~\ref{sec:annotation-qa}.

\subsection{Instruction task graph}
Instruction task graphs are DAGs, where each node represents one step in the instructions, and each edge represents the step dependency one must follow for correct assembly, as illustrated in~\autoref{fig:overview}. 
We chose this representation to accurately define step orders~\cite{peddi-etal-NEURIPS2024-captaincook4d}.
More specifically, a node can be executed once all of its prerequisite nodes (\ie nodes with incoming edges to it) have been completed.
These instruction graphs are used as an input to provide the accurate step order requirements in our benchmarking and QA annotations.
Annotation process is detailed in \S~\ref{sec:annotation-instruction-task-graph}.

\subsection{Assembly Activity}
\label{ssec:assembly-activity}
Our dataset consists of the activity of assembling take-apart toys.
We chose the activity because it possesses fundamental characteristics for assembly.
Toy assembly shares commonalities with other complex scenarios of DIY (e.g., flatpack furniture, camping tent) and industrial settings (e.g., automotive, construction): 
steps with order dependencies, parts that are rigid and decomposable, a general reversibility of steps, and use of specific tools, like screwdrivers.
We believe that these similarities, despite any other differences, make \dataset{} a meaningful initial dataset for the community to explore assembly assistants.

\subsection{Comparison with ProMQA}
\label{ssec:comaprison-with-promqa}

While \dataset{} resembles ProMQA in task formulation and QA annotation approach, there are notable differences as shown in~\autoref{tab:dataset-comparison-cooking-assembly}. 
Specifically, \dataset{} contains multiple forms of instructions, \ie parts image and task graphs, and multiple angles of the same videos. 
The fundamental difference in the activity domains (cooking vs assembly) necessitates the difference in types of actions and their target objects.
Another distinctive difference is the correction actions, which are not included in ProMQA, partially because of the irreversible nature of the cooking activity, as well as the experimental design choice. 
As a proxy for step dependency complexity, we calculate the percentage of edges between step nodes, the nodes that are not ``START'' or ``END.'' 
The ratio is 0.6 for \dataset{} and 0.8 for ProMQA. 
Additionally, there are two more differences from the QA task perspective.
The first concerns multimodality strictness, where the extent of instruction visibility is different.
While the videos in both datasets are sufficient sources for instructions, the parts images in ProMQA-Assembly are more visible compared to the textual step descriptions in ProMQA.
The second difference from the QA task perspective regards the importance of step histories, especially the last frame(s), due to the types of objects being recorded and the visibility of the components.
In ProMQA, the remaining cooking ingredients and the ingredients already added are not always captured in the last frame(s), \eg sugar melts and becomes invisible.  
In contrast, the assembled parts and the remaining parts are more visible in the last frames(s) in \dataset{}.

In short, \dataset{} is based on simpler activities, while it has more variations in terms of video views and instructions.
However, this simplicity is mainly from a human perspective, and it is unclear if the same applies to systems.
We believe that this additional domain, along with the variations that \dataset{} provides, can facilitate multi-faceted analysis to understand the capabilities of multimodal models.

\section{QA annotation}
\label{sec:annotation-qa}
We employ a semi-automatic approach for our QA annotation to ensure cost-effectiveness while maintaining quality.

\begin{figure*}[t]
    \centering
    \includegraphics[width=0.9\textwidth]{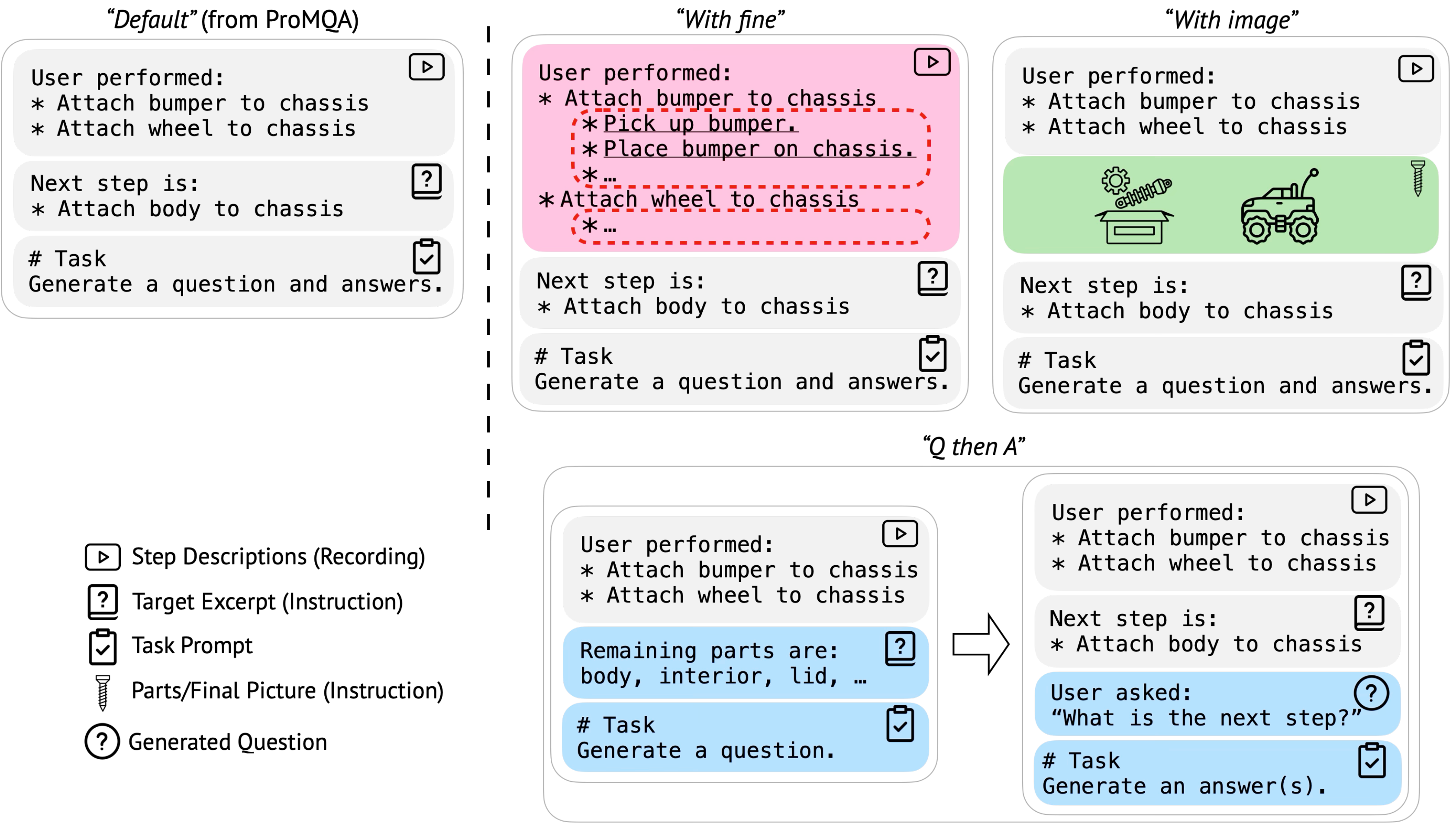}
    \caption{QA generation prompts}
    \label{fig:annnotation-prompt}
\end{figure*}

\subsection{Preprocess}
\label{ssec:qa-preprocess}
As the source of assembly recordings, we select Assembly 101~\citeplanguageresource{sener-cvpr2022-assembly101}, containing 362 unique recordings of users assembling take-apart toys. 
In the recordings, users occasionally make mistakes and need to disassemble and reassemble the parts.
This ``\textit{natural}'' behavior makes the dataset suitable for our online-style QA dataset.
We first filter the recordings based on the annotation completeness. 
Among 362 sequences, 244 sequences contain all annotations (mistake labels, coarse action labels, and fine-grained action labels) and all angles of videos (12 views). 
We also manually corrected some coarse action labels based on our manual inspections. 

Next we create candidate examples by cutting the original recordings based on the action temporal segmentation labels. 
Suppose we have a sequence with $n$ steps, $S = \{s_0, s_1, \dots, s_{n-1}\}$.
From $S$ we create $n$ video segments $v_{0:k} = \{s_0, s_1, \dots, s_{k}\}$, where $k \in \{1, 2, \dots,n\}$.
With each video segment we create eight examples by attaching a different target question type, $t$, for prompting.
The target question types reflect the four existing mistake annotations in Assembly 101, including \textbf{\textit{order}} for the wrong order, \textbf{\textit{past}} for when a previous step is a mistake, \textbf{\textit{misadjustment}} for a step that should not have happened, and \textbf{\textit{location}} for a part in the wrong position.
Additionally, we added \textbf{\textit{next}} for asking about next steps and \textbf{\textit{missing}} for asking about any missing steps, as well as \textbf{\textit{others(process)}} and \textbf{\textit{others(step)}} for general questions to increase the diversity of the questions. 
Together with our instruction task graph $g$ (\S~\ref{sec:annotation-instruction-task-graph}) and parts/target assembly image $i$ from Assembly101, one example consists of $\langle v_{0:k}, g, i, t \rangle$.
Through this process, we obtained around 22k examples in total.

Due to the time-intensive nature of manual verification, we sampled examples based on two criteria. 
First, we sample at most two examples for each target question type from one unique sequence.
Second, for the question types of \textit{order}, \textit{past}, \textit{misadjustment}, and \textit{location}, we prioritize picking the video segments where their respective mistake label is attached to the last action of each video segment. 
These conditions help us make use of as many sequences in the original data as possible and generate more relevant questions.
For all 8 question types, we sampled at most 100 examples for each type under the conditions, which resulted in 794 examples in total.

\begin{table}[!t]
\centering
\fontsize{8}{9}\selectfont
\setlength\tabcolsep{2pt}
\begin{tabular}{ccccccc}
    \toprule
    & \multicolumn{3}{c}{Prompt} & \multicolumn{3}{c}{Generator} \\
    \cmidrule(lr){2-4} \cmidrule(lr){5-7}
    Default & Q then A & w/ img & w/ fine & Gemini & Claude & o3-mini\\
    \midrule
    0.78 & 0.63 & 0.39 & 0.78 & 0.67 & 0.52 & 0.50 \\  
    \bottomrule
\end{tabular}
\caption{Prompt \& QA generator comparison. The number is the percentage of multimodal procedural questions among all generated questions.}
\label{tab:annotation-qa-generation-comparison}
\end{table}


\subsection{QA generation}
\label{ssec:qa-generation}

We used LLMs to generate candidate question-answer pairs to reduce annotation costs, following prior work~\cite{mangalam-etal-NEURIPS2023-egoschema, hasegawa-etal-2025-promqa}. 
As a starting point, we first evaluated whether ProMQA's optimal prompt was effective for our assembly domain; in addition, we tested new prompt variants under the controlled experimental settings.
We also compared four candidate QA generators.
All generated QA pairs underwent human verification to ensure quality (\S~\ref{ssec:qa-verification}).

\paragraph{Prompt comparison:}
\label{para:prompt-comparison}

\autoref{fig:annnotation-prompt} shows the abridged versions of prompt templates including \textit{``Default''}, \textit{``With fine''}, \textit{``With image''}, and \textit{``Q then A''}. 
\textit{``Default''} represents ProMQA's optimal prompt template, which mainly consists of three components: 
(1) step descriptions: textual descriptions of steps already performed to inform LLMs of the context of the questions to be generated, 
(2) target excerpt: an answer hint, extracted from instructions or mistake annotations, and
(3) task prompt: including a target question type to be generated.
New prompt variants change one aspect at a time under the controlled experiment settings.
\textit{``With fine''} is a prompt template where fine-grained step descriptions and fastening information are added on top of the existing coarse step descriptions. 
\textit{``With image''} is the prompt where we feed a parts/target assembly image as additional input to LLMs.
\textit{``Q then A''} represents two-step prompting where questions are first generated, followed by answer generation, which tries to direct LLMs to focus on each step one at a time. 
Using 46 samples, we compared these four prompt templates by generating question-answer pairs with one fixed LLM and manually checked them. 
\autoref{tab:annotation-qa-generation-comparison} shows the rate of our target multimodal procedural question (criteria in \S~\ref{ssec:qa-verification}) out of all samples. 
According to the result, \textit{``Default''} and \textit{``With fine''} generate our target questions the most. 
By comparing them qualitatively, we found that the \textit{``With fine''} template successfully integrated the fine-grained action information in the generated questions such as ``What should I do after putting down the partial toy?'' and ``Do I need to put a screw on this part?'', while generating coarse action-level questions as well. 
Considering that the \textit{``Default''} generates only coarse action-level questions, we chose the \textit{``With fine''} template in this study.

\paragraph{Model selection:}
In addition to the default GPT-4o~\cite{openai-2024-gpt4o}, we considered the following models as candidate QA generators:
Gemini 2.5 Pro~\cite{google-2025-gemini25}, Claude 3.7 Sonnet~\cite{anthropic-2025-claude37}, and o3-mini~\cite{openai-2024-o3-mini}. 
Based on the experimental results, shown in~\autoref{tab:annotation-qa-generation-comparison}, we decided to use GPT-4o as our QA generator.

\paragraph{Manual QA creation:}
In addition to automatic QA generation, we also ask a human annotator to create QA pairs for a small set of examples (50 examples) as a comparison.
The annotation took more than 12 times longer than our automatic generation (45 mins vs 3.5 mins). 
These human-created QA pairs were merged with machine-generated ones and underwent human verification. 

\begin{figure*}[t]
    \centering
    \includegraphics[width=0.88\textwidth]{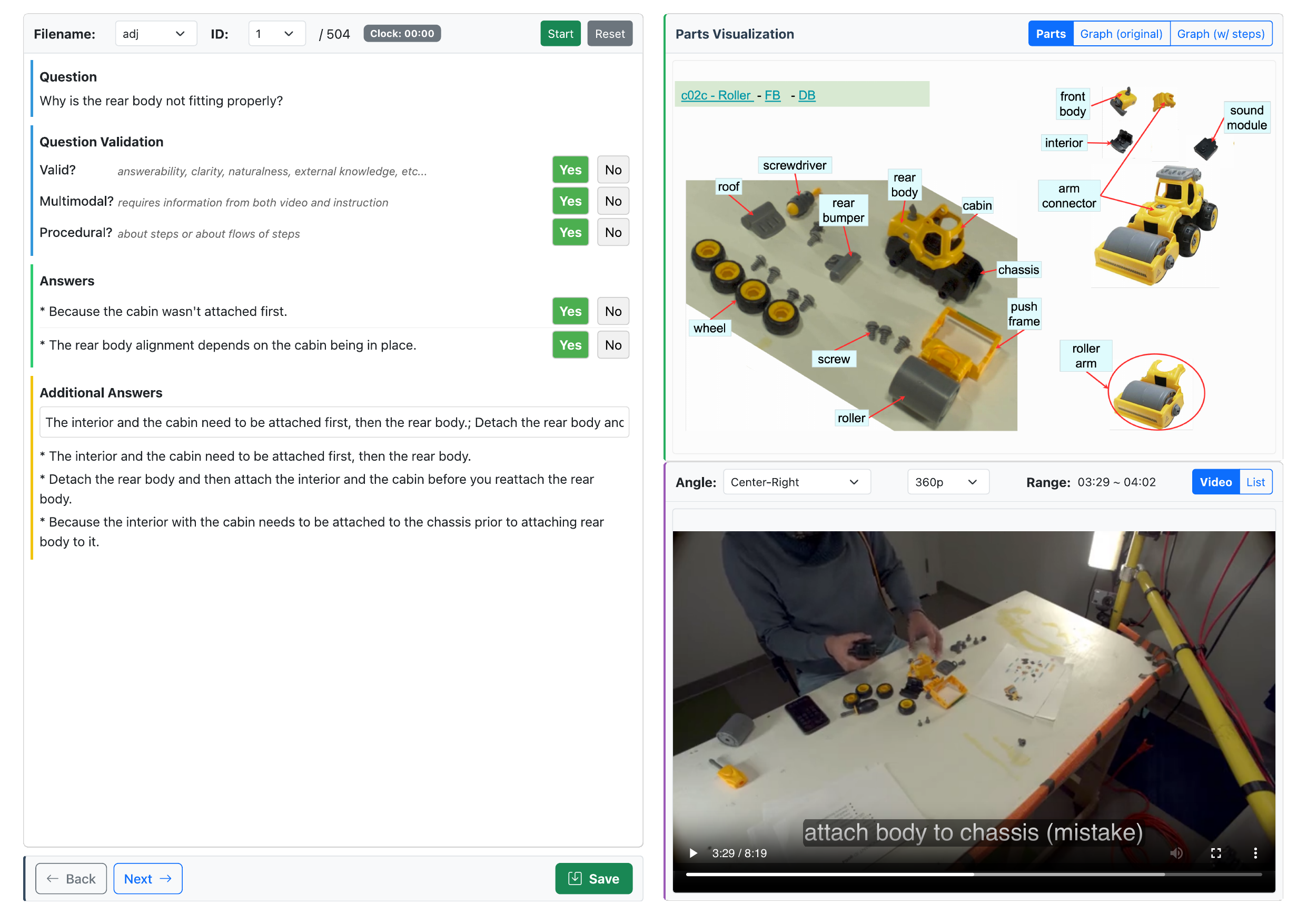}
    \caption{QA verification interface. An annotator verifies the question and answers (left panel) based on the instruction (upper right) and the recording (lower right). If needed, they write answers in the comment.}
    \label{fig:annotation-interface-qa}
\end{figure*}


\subsection{QA verification}
\label{ssec:qa-verification}
We performed human verification on \dataset{} to ensure its quality as an evaluation dataset. 
Our choice prioritizes quality over quantity, unlike automatic methods which tend to prioritize quantity over quality.
Human annotators check each QA example by referring to its corresponding instructions (both task graph and parts image) and video recording. 
Questions were verified based on three aspects: 
(1) \textbf{Valid}: whether or not a question is answerable and makes sense. 
If not, annotators can skip to the next example, 
(2) \textbf{Multimodal}: whether or not a question requires information from both its instructions and recording, and 
(3) \textbf{Procedural}: whether or not a question is about a step or a sequence of steps.
Answers were simply checked to see whether they were correct or not. 
When candidate answers were incorrect or incomplete, annotators provided the correct/missing answer(s).
For each example, we assign two annotators to independently verify each QA example. 
When two annotations conflict in any of the aspects, another annotator acts as an adjudicator to resolve any annotation conflicts. 
After a trial annotation, five annotators and one adjudicator performed the verification task using the annotation guideline developed during the trial.
All annotators and the adjudicator have CS-related graduate degrees.
We calculated the inter-annotator agreement (IAA) based on the percent agreement: $0.75$ for the valid aspect, $0.83$ for the multimodal aspect, $0.96$ for the procedural aspect, and $0.77$ for the answer correctness.
After the adjudication, we were left with 646 valid questions, of which 538 were multimodal procedural questions, and 47 were human-created questions.

\section{Instruction task graph annotation}
\label{sec:annotation-instruction-task-graph}

\begin{figure*}[t]
    \centering
    \includegraphics[width=0.87\textwidth]{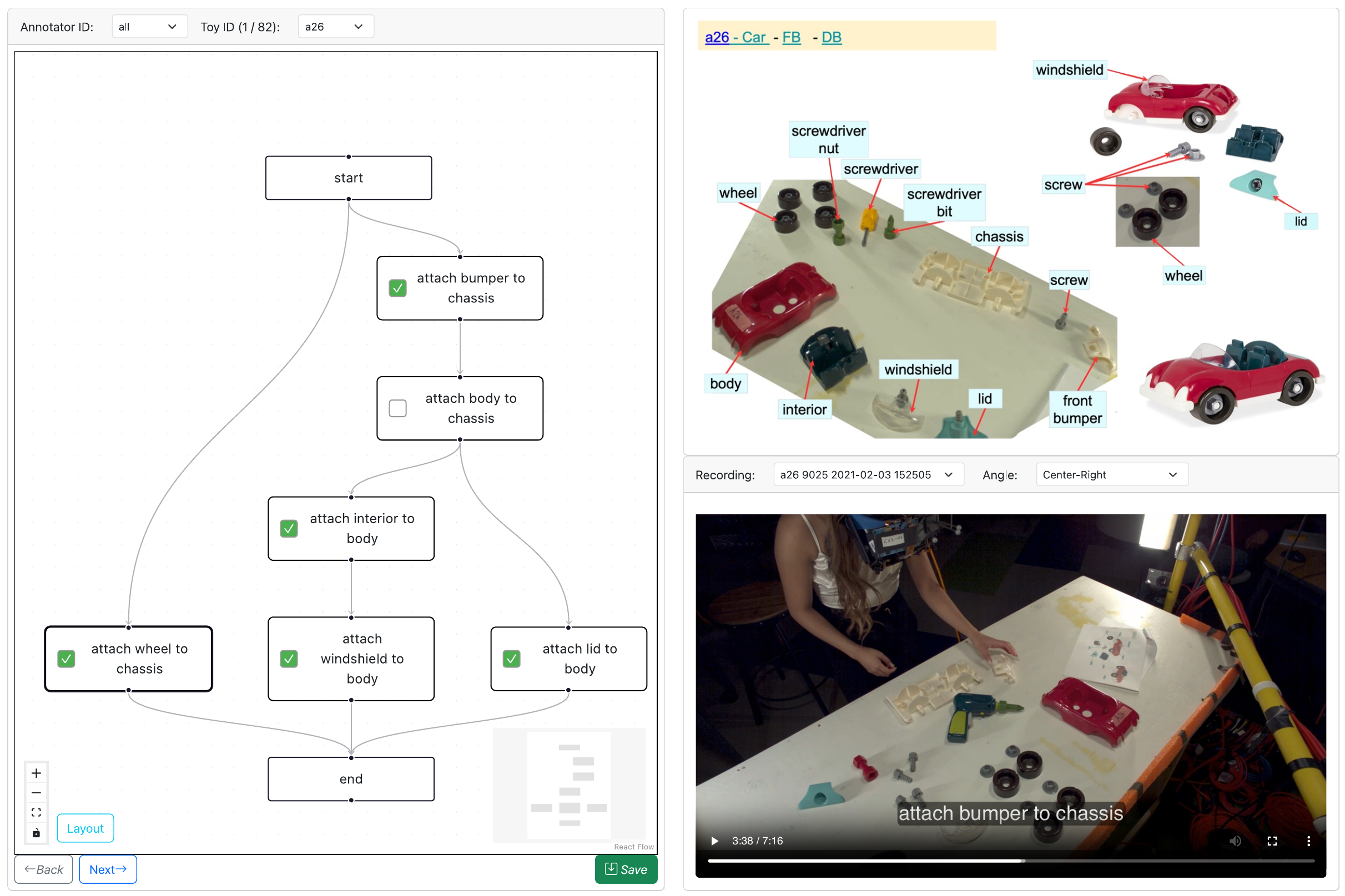}
    \caption{Task graph annotation interface. An annotator checks the part image (upper right) and the recordings (lower right) to identify the step dependencies and add edges by dragging on the graph annotation interface (left).}
    \label{fig:annotation-interface-graph}
\end{figure*}

In addition to QA pairs, we annotated instruction task graphs for the toys in Assembly101 motivated by three reasons. 
First, the task graphs represent accurate step order requirements, which are essential in benchmarking experiments. 
Second, the graphs are used in our QA generation process (\S~\ref{ssec:qa-generation}), specifically to provide answer hints as target excerpts.
Third, they can facilitate the human verification process. 
During our preliminary QA verification, we found the task challenging if we could only watch the recording and look at the parts' image for the question, especially when recall is important. 
For instance, ``What is the next step now?'' should have all possible next steps as answers, and even humans may miss listing all the next steps if they are unable to refer to a verified task graph. 


\subsection{Annotation procedure}
\autoref{fig:annotation-interface-graph} shows the annotation interface.
The initial sets of nodes were collected based on the coarse action labels in Assembly101, with two extra nodes, ``START'' and ``END,'' for each graph. 
Annotators can watch a set of recordings of multiple users assembling the same toy, often in different orders. 

Given the pre-loaded nodes and a set of recordings, annotators add edges between nodes by identifying a specific order of steps to assemble the toy correctly.
Additionally, we ask annotators to check if a step requires a fastening (\ie a screw) to provide further enriched step information in QA generation. 
We assign two people to annotate each graph independently, and if any conflicts happen, an adjudicator resolves them. 
Five people participated as annotators, and one person served as an adjudicator.
In total, we annotated 81 instruction task graphs, which is three times more than the task graphs in CaptainCook4D.  
To measure IAA, we calculated pairwise F1 scores~\cite{uzzaman-allen-2011-temporal} for the edge labeling and percent agreements for fastening checks. 
The average IAA for edges is $0.74$, and that for fastening is $0.89$. 
While these manually curated instruction task graphs are used as inputs in this study, they can be used to evaluate systems for task graph construction as well~\cite{seminara-neurips2024-differentiable}. 
We encourage the researcher to test their graph construction systems on our 81 task graphs, in addition to the 24 graphs in CaptainCook4D. 
\section{Benchmarking}
\label{sec:benchmarking}

We provide benchmarking results on \dataset{} to give insights to the research community on how well existing models perform.

\begin{table*}[!t]
\centering
\fontsize{8}{9}\selectfont
\setlength\tabcolsep{4pt}
\begin{tabular}{l c cc cccccc cc cc}
    \toprule
    \multirow{2}{*}[-1ex]{Model} & \multirow{2}{*}[-1ex]{All} & \multicolumn{2}{c}{Multi\&Proc} & \multicolumn{6}{c}{Process-level} & \multicolumn{2}{c}{Step-specific} & \multicolumn{2}{c}{Source} \\
    \cmidrule(lr){3-4} \cmidrule(lr){5-10} \cmidrule(lr){11-12} \cmidrule(lr){13-14}
    & & Yes & No & next & miss & order & misadj & past & \makecell[c]{others} & loc & \makecell[c]{others} & ma & hum \\
    \midrule
    \underline{Text-only}\\
    ~~Llama 3.3 72B & 30.6 & 27.6 & 45.4 & 39.5 & 36.0 & 43.3 & 29.5 & 29.4 & 27.3 & 13.4 & 31.6 & 30.5 & 31.9 \\
    ~~DeepSeek R1 (Qwen 32B) & 44.0 & 39.6 & 65.7 & 47.0 & 51.9 & 60.8 & 42.9 & 30.4 & 27.3 & 34.6 & 48.0 & 44.1 & 42.6 \\
    ~~Qwen2.5-VL & 39.7 & 37.5 & 50.9 & 48.0 & 57.5 & 42.5 & 39.3 & 31.4 & 31.8 & 19.1 & 46.9 & 39.4 & 43.6 \\
    ~~GPT-4o & 34.4 & 31.1 & 50.9 & 40.5 & 46.7 & 49.2 & 33.9 & 29.4 & 22.7 & 15.1 & 40.8 & 34.7 & 30.9 \\
    ~~GPT-5 w/ think & 47.1 & 42.1 & 72.2 & 49.7 & 63.1 & 65.0 & 44.6 & 33.3 & 31.8 & 29.2 & 61.2 & 47.2 & 46.8 \\
    ~~Claude 3.7 Sonnet & 36.4 & 31.5 & 60.6 & 45.1 & 45.3 & 50.0 & 38.4 & 37.3 & 29.5 & 13.8 & 41.8 & 36.3 & 37.2 \\
    ~~Gemini 2.5 Pro & 36.3 & 30.8 & 63.9 & 45.7 & 47.7 & 62.5 & 31.2 & 21.6 & 25.0 & 15.8 & 38.8 & 35.9 & 41.5 \\
    \cmidrule(lr){1-14}
    \underline{Multimodal} \\
    ~~Qwen2.5-VL 72B Instruct & 45.4 & 44.4 & 50.0 & 46.1 & 53.7 & 40.0 & 42.9 & 39.2 & 38.6 & 43.0 & 51.0 & 44.1 & 61.7 \\
    ~~LLaVA-Video 72B (Qwen2) & 47.4 & 46.5 & 51.9 & 47.0 & 57.5 & 31.7 & 40.2 & 50.0 & 31.8 & 49.3 & 52.0 & 47.1 & 51.1 \\
    ~~InternVL3 8B & 38.2 & 37.2 & 43.5 & 38.5 & 27.6 & 41.7 & 34.8 & 36.3 & 25.0 & 45.0 & 48.0 & 37.7 & 44.7 \\
    ~~GPT-4o & 47.8 & 46.5 & 54.2 & 44.1 & 43.5 & 59.2 & 45.5 & 49.0 & 36.4 & 50.3 & 53.1 & 48.1 & 43.6 \\
    ~~GPT-4o w/ CoT & 45.8 & 41.1 & 69.4 & 48.0 & 41.6 & 65.8 & 42.9 & 37.3 & 38.6 & 41.9 & 51.0 & 45.2 & 54.3 \\
    ~~GPT-5 & 48.2 & 46.3 & 57.9 & 48.4 & 41.1 & 56.7 & 41.1 & 44.1 & 34.1 & 53.4 & 56.1 & 47.7 & 55.3 \\
    ~~GPT-5 w/ think & \textbf{58.0} & \textbf{53.2} & \textbf{81.9} & 53.3 & \textbf{65.9} & \textbf{72.5} & 48.2 & 48.0 & 38.6 & \textbf{55.7} & \textbf{74.5} & \textbf{58.2} & 55.3 \\
    ~~Claude 3.7 Sonnet & 47.8 & 45.2 & 60.1 & 52.9 & 43.5 & 63.1 & 36.0 & \textbf{55.8} & 32.6 & 42.0 & 45.0 & 47.4 & 51.0\\
    ~~Claude 3.7 Sonnet w/ CoT & 48.4 & 43.8 & 71.3 & 51.3 & 52.3 & 65.8 & 40.2 & 49.0 & 31.8 & 39.9 & 51.0 & 48.2 & 50.0 \\
    ~~Claude 3.7 Sonnet w/ think & 48.5 & 44.1 & 69.9 & 53.3 & 51.9 & 66.7 & 39.3 & 46.1 & 29.5 & 40.9 & 48.0 & 47.9 & 55.3 \\
    ~~Gemini 2.5 Pro & 51.0 & 45.4 & 78.7 & 56.6 & 59.3 & 66.7 & 39.3 & 34.3 & 18.2 & 46.3 & 56.1 & 50.1 & 62.8\\
    ~~Gemini 2.5 Pro w/ CoT & 54.2 & 49.5 & 77.3 & \textbf{58.9} & 60.7 & \textbf{72.5} & \textbf{49.1} & 42.2 & \textbf{40.9} & 44.0 & 58.2 & 53.3 & \textbf{66.0} \\
    \cmidrule(lr){1-14}
    ~~Human* & (70.7) & --- & --- & --- & --- & --- & --- & --- & --- & --- & --- & --- & --- \\
    \bottomrule
\end{tabular}
\caption{Benchmarking result. ``Multi\&Proc'' represents multimodal procedural questions. Categories under ``Process-level'' and ``Step-specific'' are question types. ``Source'' shows the creators of questions, either machine (``ma'') or human (``hum''). All numbers are averages over examples for each category. *: Human performance is based on the sampled 20 examples.}
\label{tab:benchmark-result}
\end{table*}

\subsection{Experimental Setup}
We consider three types of models.
One is a text-only model, where we feed only an instruction task graph and a question without any vision inputs to obtain the lower bound for multimodal-input models. 
We tested Llama 3.3 (72B)~\citep{grattafiori-arxiv2024-llama3} and DeepSeek-R1~\cite{deepseekai-2025-deepseekr1}.
For the DeepSeek-R1, we used the distilled version with Qwen (32B), due to our computational resources. 
Next, we chose open(-weight) multimodal models including Qwen2.5-VL (72B)~\cite{bai-arxiv2025-qwen25-vl}, a smaller version of InternVL3 (8B)~\cite{zhu-arxiv2025-internvl3}, and LLaVA-Video (72B)~\cite{zhang-arxiv2024-video}.
Finally, we also benchmarked proprietary multimodal models, namely, GPT-4o~\cite{openai-2024-gpt4o}, GPT-5~\citep{openai-2025-gpt5}, Gemini 2.5 Pro~\cite{google-2025-gemini25}, and Claude 3.7 Sonnet~\cite{anthropic-2025-claude37}, with zero-shot chain-of-thought (\textit{CoT})~\citep{kojima-neurips2022-large} or their reasoning modes (\textit{think}).
Additionally, we evaluated some multimodal models under the text-only setting to highlight the importance of the vision and multimodal inputs. 
As a comparison, we also obtained human performance by asking three of the annotation participants to answer questions that they had not checked in the verification stage.
Due to cost constraints, we used only a subset of 20 different questions out of the full 646 questions.

The outputs of all open-weight models are obtained using at most four A6000 GPUs.
We sample 20 frames uniformly from each recording from the above angle with resizing to 640x360, and feed them with a corresponding text prompt and a parts image.
Task graphs are fed in DOT format.
For evaluation, we use LLM-as-a-judge~\cite{zheng-neurips2023-judging} with GPT-4o, following ProMQA. 
The evaluator provides scores in a three-point Likert scale, and we scale them to 0-100 by multiplying by 50.

\subsection{Result}
\autoref{tab:benchmark-result} shows the benchmarking result. 
Based on this result, we focused on two points: the confirmation of our dataset's value and a promising approach with room for improvement. 
As for the former point, models perform consistently better under the multimodal setting than under the text-only setting, which supports the necessity of visual understanding for \dataset{}. 
For instance, the same proprietary models perform even more than 10 points better with visual inputs compared to text-only settings ($51.0$ with multimodal vs $36.3$ with text-only settings for Gemini 2.5 Pro).
In addition, the general performance gap between multimodal procedural understanding questions (\textit{Yes}) and other valid questions (\textit{No}) strengthens the value of the challenge that our dataset poses.
Specifically, GPT-5 (\textit{w/ think}) answers our valid, yet non multimodal-procedural questions ($81.9$) much easily, compared to our target questions ($53.2$).
Concerning the second point, models with reasoning capability stand out in each setting, i.e., DeepSeek-R1, GPT-5 (\textit{w/ think}), and Gemini 2.5 Pro (\textit{w/ CoT}), among models with mostly comparable performance.
This indicates the potential of inference-time scaling methods and reasoning-oriented training.
While GPT-5 (\textit{w/ think}) shows the best performance amongst the ones we tested, it still lags far behind the reference human performance.
Our benchmark experiments show promising results of reasoning techniques, and we would encourage the research community to explore more to further advance the multimodal procedural activity understanding by utilizing our \dataset{} dataset.
\section{Conclusion}
\label{sec:conclusion}

In this work, we propose a new multimodal QA evaluation dataset, \dataset{}, which specifically targets assembly tasks.
During the development phase, we annotate QA pairs efficiently using the combination of LLM generation and human verification, and we also create instruction task graphs.
Our benchmarking results show that current multimodal models underperform humans, which suggests great room for improvement.
We believe that our benchmark dataset sheds light on the missing capabilities of multimodal models and facilitates the development of further helpful systems for humans.

\clearpage

\section{Ethical Statement}
In our dataset development we used LLMs, which are pretrained on a massive web-scraped corpora. 
Considering that LLMs may introduce prejudiced, offensive, or biased content in our dataset, we carefully checked potentially inappropriate questions and answers in our verification process.

\section{Limitations}
We acknowledge several limitations of our work. 
First, the size of our dataset is relatively small. 
This is based on our design choice of quality over quantity, due to the careful curation by the human verification.
While our dataset can be reliably used to evaluate models, which is the primary purpose of this dataset, when it comes to training models, one may need to resort to automatic verification with LLMs.
Since our approach produces 22k examples in total at the preprocessing time, effective automatic verification has the potential to scale the dataset.

The second point concerns real-world applicability due to the target objects of take-apart toys.
Take-apart toy assembly may be simpler for humans than other practical assembly tasks.
However, as we discussed in \S~\ref{ssec:assembly-activity}, take-apart toys possess fundamentally common characteristics with broader assembly tasks. 
Hence, the findings based on our dataset, together with our annotation approach, can be generalized to broader scenarios.
At the same time, developing QA datasets on higher complexity tasks is beneficial for further practicality, although few resources in the current research community contain error-included real recordings of assembly tasks. 
Furthermore, much greater variability exists in real-world use cases, \eg insufficient light, vision obstacles, third-person interactions, evolving to a dialog-style setting, and/or replacing the task graphs with normal assembly instructions, which requires the identification of implicit step dependencies~\cite {lal-etal-2024-cat}.
We envision our work encouraging the community to invest in resource development on higher-complexity assembly tasks in more practical environments.

Third, the dataset contains only English.
With the increase of multilingual multimodal models, the English capability on procedural activity understanding can be indicative of the capabilities with other languages on similar tasks.
Yet, considering the ubiquitous demand for assembly assistant systems, we hope to contribute to developing datasets in non-English languages.
Finally, we note that our dataset is released for evaluation purposes only, not for training, in compliance with OpenAI's terms of use.\footnote{https://openai.com/policies/row-terms-of-use/}

\section{Acknowledgment}
This work is partially supported by (1) Programs for Bridging the gap between R\&D and the IDeal society (society 5.0) and Generating Economic and social value (BRIDGE) / Practical Global Research in the AI $\times$ Robotics Services, implemented by the Cabinet Office, Government of Japan, and (2) a project, JPNP25006, commissioned by the New Energy and Industrial Technology Development Organization (NEDO).

\section{Bibliographical References}
\label{sec:reference}
\bibliographystyle{lrec2026-natbib}
\bibliography{reference}


\clearpage
\appendix
\section{QA annotation}

\subsection{Preprocess}
To make full use of fine-grained action labels, the preprocessing for \textit{``With fine''} is slightly different from others in terms of video cutting points.
For \textit{``With fine''}, we first cut a recording into video segments based on coarse actions, then we slightly move the end timestamp backward based on the fine action labels to add diversity in end points. 
Specifically, suppose the latest step $s_k$ in a video segment $v_{0:k}$ contains $l$ fine-grained steps, $s_k = \{s_{k,0}, \dots, s_{k,l-1}\}$. 
We randomly choose one fine-grained step $s_{k,j}$, and move the end timestamp.

As for manual coarse-action correction, one annotator manually skimmed through and corrected the labels. 
Representative modifications include typos, normalizing steps descriptions (\eg ``attach interior to chassis'' in one recording and ``attach chassis to interior'' in another recording for the same toy), and misspecification (\eg ``attach body to base,'' where there are ``front body'' and ``rear body'').
On average, 1-2 corrections were made for each toy. 
We release the modifications, in addition to other data, for reproducibility. 

\subsection{QA generation}

In our prompt comparison experiment, we compared \textit{``Default''}, \textit{``With fine''}, \textit{``With image''}, and \textit{``Q then A''}. 
The full prompt examples of \textit{``Default''}, \textit{``With fine''}, and \textit{``With image''} are shown in~\autoref{fig:prompt-full-qa-generation-default},~\ref{fig:prompt-full-qa-generation-with-fine}, and ~\ref{fig:prompt-full-qa-generation-with-image}, while~\autoref{fig:prompt-full-qa-generation-q} and~\ref{fig:prompt-full-qa-generation-a} present the full prompt examples for generating questions and answers for \textit{``Q then A''}, respectively.

In the qualitative comparison between \textit{``Default''} and \textit{``With fine''}, we found that \textit{``With fine''} generated 25 questions (out of 46) that mentioned fine-grained step information, while 2 questions by \textit{``Default''} mentioned fine-level step information.
Furthermore, we calculated the distinct-1, \ie type-token ratio, and distinct-2~\cite{li-etal-2016-diversity} to measure the lexical diversity.
We first removed stop words and calculated these scores. 
The \textit{``Default''} showed distinct-1 and distinct-2 scores of $0.30$ and $0.77$, respectively, while \textit{``With fine''} improved these metrics to $0.36$ and $0.80$. 

Our manual investigation of the \textit{``Q then A''} results revealed that this template generated ``should'' questions more frequently than the above two templates. These questions tend to require only the instruction information in most cases (\eg ``Should I attach the roof before the light?'').
While the template may allow a model to focus on each stage of generation one by one, knowing the target answer in the question generation stage may help a model to generate a question that requires both what has been done and what is supposed to be done to answer.
Additionally, we found that \textit{``Q then A''} generated more incorrect answers, which would increase the burden on annotators. 
One typical reason was because a model generated a question that was different from the specified type in its prompt, hence, the provided target excerpt, \ie answer hint, became noise for the answer generation. 
Suppose a prompt type is ``missing,'' where ``attach body to chassis'' needs to be done before the latest action ``attach interior to chassis.''
In this case, the target excerpt is the missing steps, ``attach body to chassis.''
However, if the generated question is about next steps, \eg ``What should I do next?'', the target excerpt misleads the answer generation. 
In this case, the correct answer would be ``You should detach the interior first and attach the body first before the interior,'' but the model is likely to generate ``The next step is to attach the body to the chassis.''
Moreover, in the \textit{``Q then A''} setting, a model often generated both yes and no answers for the same question and/or paraphrases of the same answers, presumably because the model tried to increase the recall. 
As we keep only the correct answers, this recall-oriented behavior increases the workload in the verification stage. 

As for \textit{``With image''}, we suspect that the input image is distracting for a model, rather than helpful, possibly due to its tendency to draw high attention to irrelevant visual tokens~\cite{leng-cvpr2024-mitigating}. 

In our preliminary experiments, we used the following versions of the models: 
GPT-4o (\texttt{gpt-4o-2024-11-20}), Gemini 2.5 Pro (\texttt{gemini-2.5-pro-exp-03-25}), Claude 3.7 Sonnet (\texttt{claude-3-7-sonnet-20250219}), and o3-mini (\texttt{o3-mini-2025-01-31}). 

In the manual QA creation, we provide a similar interface to that for QA verification, specifically, a recording with a textual action list and a part image. 
The annotator writes a question, its answer(s), and the timestamp for the question to be asked. 
The annotation guideline for the manual QA creation contains the interface description and the explanation of our target multimodal procedural questions, which are similar to what a model receives in QA generation. 
One difference is that we did not specify question types so that the annotator can decide on a question type depending on the context.

\subsection{QA verification}

\begin{figure*}[t]
    \centering
    \includegraphics[width=0.7\textwidth]{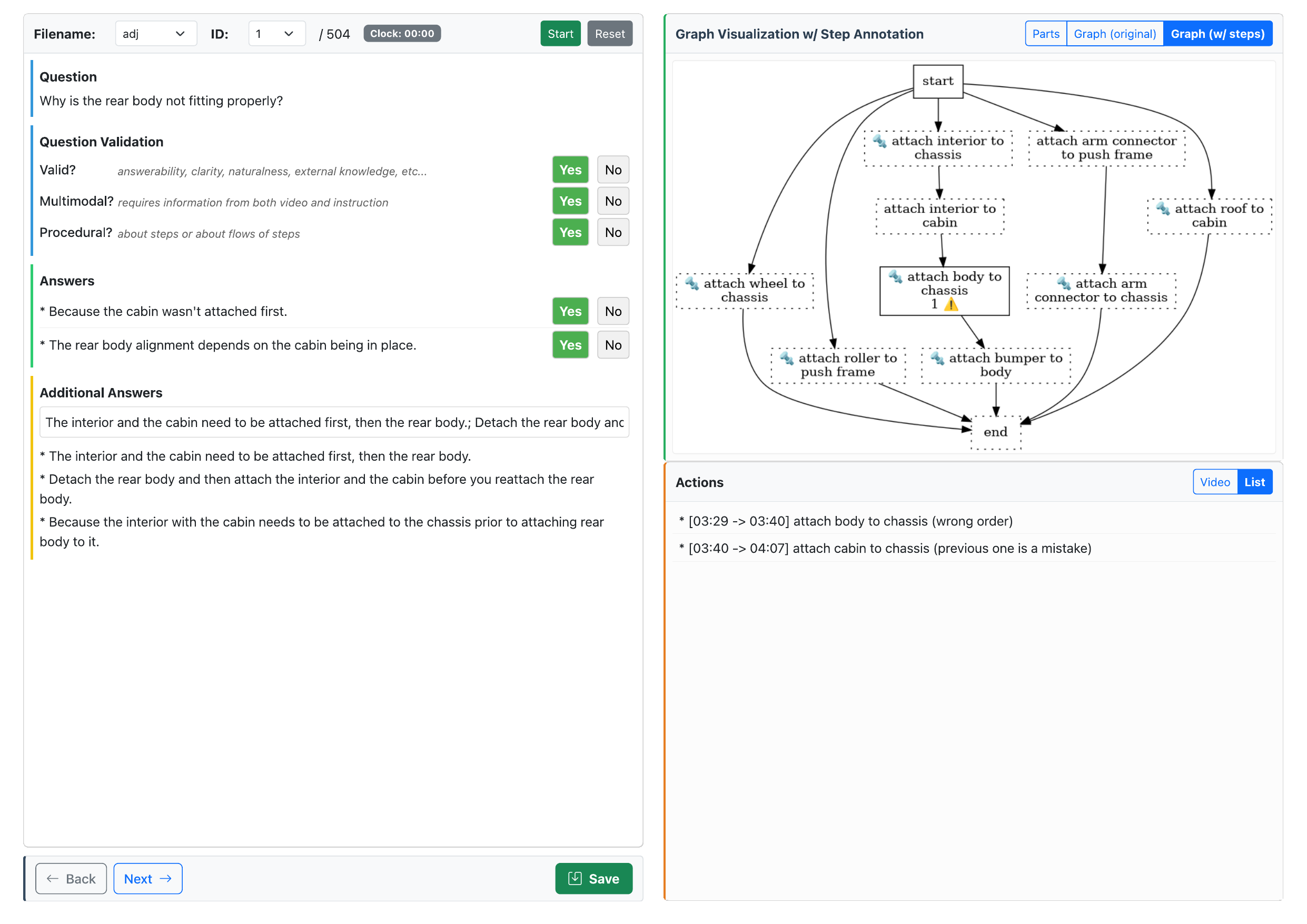}
    \caption{The different view of QA verification interface.}
    \label{fig:annotation-interface-qa-2}
\end{figure*}

\autoref{fig:annotation-interface-qa-2} shows another view of our QA verification interface: the task graph with step information and the textual list. 

Considering that the human annotator who created QAs manually is one of the annotators in the verification stage, we split the manually-created QAs among the annotators who did not create QAs. 

We select the adjudicator among the annotators based on their familiarity with the task. 
The adjudicator is the one who participated in the development of our guideline. 
All annotators read the guideline first and annotate 10 examples, followed by our feedback, before completing the remaining ones. 

Among the 646 questions, 599 were generated by LLMs, and 47 were created by the human annotator. 
IAA for ``Multimodal,'' ``Procedural,'' and answers are calculated only for questions that are marked as valid by two annotators.

\section{Instruction task graph annotation}

The IAA for edges is calculated as a pairwise F1 score.
The recall and the precision are defined as follows~\cite{uzzaman-allen-2011-temporal}:
\begin{equation*}
    \text{Precision} = \frac{|E_1 \cap E_2|}{|E_1|}, \hspace{0.2cm}
    \text{Recall} = \frac{|E_1 \cap E_2|}{|E_2|}
\end{equation*}
, where $E_1$ and $E_2$ denote the sets of edges from two independently annotated task graphs.
The guideline for the task graph annotation is also included in our supplemental material. 

Four of the five annotators were the same for both the task-graph annotation and QA verification, with one unique annotator assigned to each task, due to their availability.


\section{Benchmarking}

\subsection{Experimental setup}

In our benchmarking experiment, we used the following versions of the models:
DeepSeek R1 (\texttt{deepseek-ai/DeepSeek-R1-Distill-Qwen-32B}), Qwen2.5-VL (\texttt{Qwen/Qwen2.5-VL-72B-Instruct}), LLaVA-Video (\texttt{lmms-lab/LLaVA-Video-72B-Qwen2}), InternVL3 (\texttt{OpenGVLab/InternVL3-8B}), GPT-4o (\texttt{gpt-4o-2024-11-20}), GPT-5 (\texttt{gpt-5-2025-08-07}), Claude 3.7 Sonnet (\texttt{claude-3-7-sonnet-20250219}), and Gemini 2.0 Flash (\texttt{gemini-2.0-flash-001}).
Proprietary models are based on their API services, and open-weight models are downloaded from the HuggingFace Hub\footnote{\url{https://huggingface.co/}} and run locally.
For the reasoning mode, we use \texttt{medium} for GPT-5 and $4098$ as budget tokens for Claude 3.7 Sonnet. 
For the DeepSeek model, we used the temperature $0.6$ based on the recommendation by the official repository.

We feed task graphs as text by representing them in DOT format.\footnote{\url{https://en.wikipedia.org/wiki/DOT_(graph_description_language)}}
In the frame sampling, we sampled uniformly from the last so that the last frame is always included. 
This is based on our insight that the last frame typically contains informative information for the QA task. 
\autoref{fig:prompt-full-benchmark} shows the basic example of the prompt in benchmarking. 
We slightly adapted the prompt for each model. 
Refer to our code attached for the details.
It took up to a few hours to run the models, and the APIs cost less than 1 USD for each.
\autoref{fig:prompt-full-benchmark-eval} shows an example prompt for our LLM-as-a-judge evaluation.

\subsection{Result}

In the human performance experiment, a participant is given a question, instructions (parts' image and task graph), and a recording (video segments, instead of sampled frames), and answers the question.
We asked three participants to answer questions, followed by the same LLM-as-a-judge, and then took the average. 
Due to the cost, we used only the subset of 20 questions out of the full 391 questions.
As the annotators are the ones who performed the QA verification, we sampled different sets of 20 questions for each so that participants answered questions that they did not see in the verification stage.
The average score was $67.5$.
Based on our manual inspection, humans make mistakes typically by failing on step status tracking and overlooking small, yet salient visual details. 
While the task is even difficult for humans who are familiar with the task, there is a gap between the human average score and the top-performing model's score, which shows room for improvement for models. 

\subsection{Qualitative Analysis}

\begin{table}[!t]
\centering
\caption{Bias analysis result: generator-predictor}
\setlength\tabcolsep{4pt}
\begin{tabular}{l ccc}
    \toprule
    \multirow{2}{*}{Predictor} & \multicolumn{3}{c}{Generator} \\
    \cmidrule(lr){2-4}
    & GPT & Claude & Gemini \\
    \midrule
    GPT & 47.5 & 42.5 & 47.5 \\
    Claude & 35.0 & 27.5 & 37.5 \\
    Gemini & 27.5 & 30.0 & 37.5 \\
    \bottomrule
\end{tabular}
\label{tab:analysis-bias-gen-pre}
\end{table}
\begin{table}[!t]
\centering
\caption{Bias analysis result: predictor-evaluator}
\setlength\tabcolsep{4pt}
\begin{tabular}{l ccc c}
    \toprule
    \multirow{2}{*}{Predictor} & \multicolumn{3}{c}{Evaluator} & \multirow{2}{*}{Human}\\
    \cmidrule(lr){2-4}
    & GPT & Claude & Gemini & \\
    \midrule
    GPT & 44.2 & 40.0 & 50.8 & 45.8 \\
    Claude & 33.3 & 30.8 & 39.2 & 33.3 \\
    Gemini & 30.8 & 30.8 & 36.7 & 31.7 \\
    \bottomrule
\end{tabular}
\label{tab:analysis-bias-pre-eval}
\end{table}

\autoref{tab:qualitative-example1} and~\ref{tab:qualitative-example2} show predictions from some of the models we evaluated.
One typical failure case is that models pay more attention to instructions when frames also contain critical information.
Interestingly, we found that the text-only model, DeepSeek-R1, sometimes predicted, or more precisely, \textit{guessed}, correctly, only based on instructions, even though its reasoning process explicitly stated that it did not have access to the information of what the user was doing.

\subsection{Bias Analysis}

We also conducted self-preference bias analysis.
Self-preference bias is a bias that LLMs favor their outputs over others~\cite{panickssery-neurips2024-llm-evaluator}.
Since our work employs LLMs in multiple places, this type of bias may unexpectedly affect the benchmark results.
As conducted in ProMQA, we investigate two types of biases to justify the quality of \dataset{} as a benchmark dataset.
One is the bias between a QA generator and a predictor. 
Minor styles or characteristics of questions generated by a model may make it easier for the same model to predict the answers. 
The other is the bias between a predictor and an evaluator.
This is the same setting as the original self-preference bias. 

We first sampled a new set of 40 samples obtained in \S~\ref{ssec:qa-preprocess}, and prompted three generators (GPT4o, Claude 3.7 Sonnet, and Gemini 2.0 Flash) to generate QA pairs. 
Due to cost constraints, we used only one annotator to verify the questions and, based on the verification results, we obtained 20 valid multimodal procedural questions from each generator.
When there were more than 20 target questions, we sampled from them.
Next, we use the same set of LLMs to predict answers to each set of 20 questions, followed by manual judgment.
This manual judgment used the same guidelines as our LLM-as-a-judge and was conducted by one annotator. 
According to the result in~\autoref{tab:analysis-bias-gen-pre}, we do not see an indication of the generator-predictor self-preference bias here. 
For instance, GPT-4o (predictor) performs the best on the GPT-4o-generated data compared to other predictors, but the trend is the same for the generated data by other models.
Or, Gemini (predictor) performed the best on its data, but that is the same for other predictors. 

Next, we checked the predictor-evaluator self-preference bias. 
We prompted the same set of LLMs to evaluate the predicted answers we obtained above, \ie $20~\text{QAs} \times 3~\text{generator} \times 3~\text{predictor} = 180~\text{predictions}$.
The result is shown in~\autoref{tab:analysis-bias-pre-eval}, as well as the human evaluation results. 
GPT-4o (evaluator) gave the best score to GPT-4o (predictor), but other evaluators consistently gave the best score to GPT-4o (predictor). 
Gemini (predictor) received the best score from Gemini (evaluator). Again, however, other models received the highest scores from Gemini (evaluator).
Based on this result, we do not see the sign of the predictor-evaluator self-preference bias, either.
In this bias analysis, the person who served as an adjudicator in \S~\ref{ssec:qa-verification} performed the verification and the manual evaluation.
We also calculated the judgment correlations between each model and the human evaluator using Kendall's Tau coefficient~\cite{kendall-1938-new}. 
The correlations were 0.26 for GPT-4o, 0.27 for Claude 3.5 Sonnet, and 0.24 for Gemini 2.0 Flash. 
While we used the evaluation prompt and GPT-4o based on ProMQA's results, there remains room for improvement in evaluation methodology, which we leave for future work.\footnote{Icons are from Freepik, \url{www.flaticon.com}}



    

\begin{table*}[!t]
\centering
\caption{
    Qualitative Analysis Example. The question requires visual understanding to identify that all the parts have been attached. While Qwen2.5-VL correctly answers, other models fail to answer. One possible reason for GPT-4o and Claude 3.7 Sonnet is that they rely too much on the instructions, instead of the frames, to answer the question, based on their answers. 
}
\setlength\tabcolsep{4pt}
\begin{tabular}{l p{10cm}}
    \toprule
    \multicolumn{2}{c}{\includegraphics[width=0.95\linewidth]{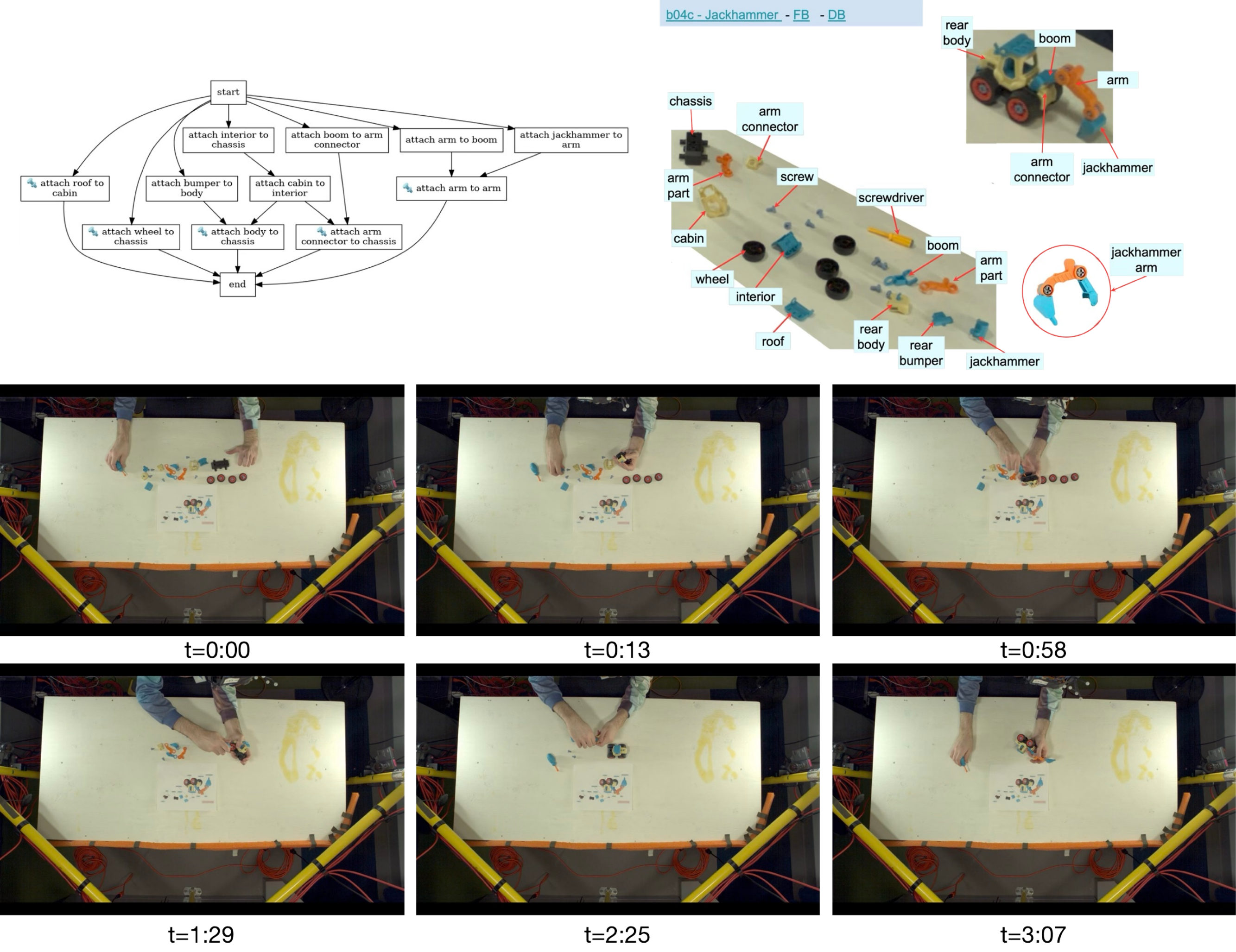}}\\
    Question: & Am I done, or is there something left to finish? \\
    Gold answer: & You're done! The toy is complete. \\
    \midrule
    DeepSeek-R1: & \xmark~There are still steps left to complete. \\
    Qwen2.5-VL: & \cmark~You are done. \\
    GPT-4o: & \xmark~You are not done yet; you still need to attach the jackhammer to the arm and complete the remaining steps as per the manual. \\
    Claude 3.7: & \xmark~Based on the assembly recording and the manual, you are not done yet. There are still several steps to complete, including attaching the jackhammer assembly to the main vehicle and connecting various components like the roof to the cabin with screws. \\
    \bottomrule
\end{tabular}
\label{tab:qualitative-example1}
\end{table*}

\begin{table*}[!t]
\centering
\caption{
    Qualitative Analysis Example. All multimodal models fail to identify what the user is doing, \ie they forgot to secure the interior to the chassis, so they are detaching the cabin. While this question requires visual understanding, DeepSeek guesses somewhat correctly, presumably because attaching the interior is one of the prerequisite steps.
}
\setlength\tabcolsep{4pt}
\begin{tabular}{l p{10cm}}
    \toprule
    \multicolumn{2}{c}{\includegraphics[width=0.95\linewidth]{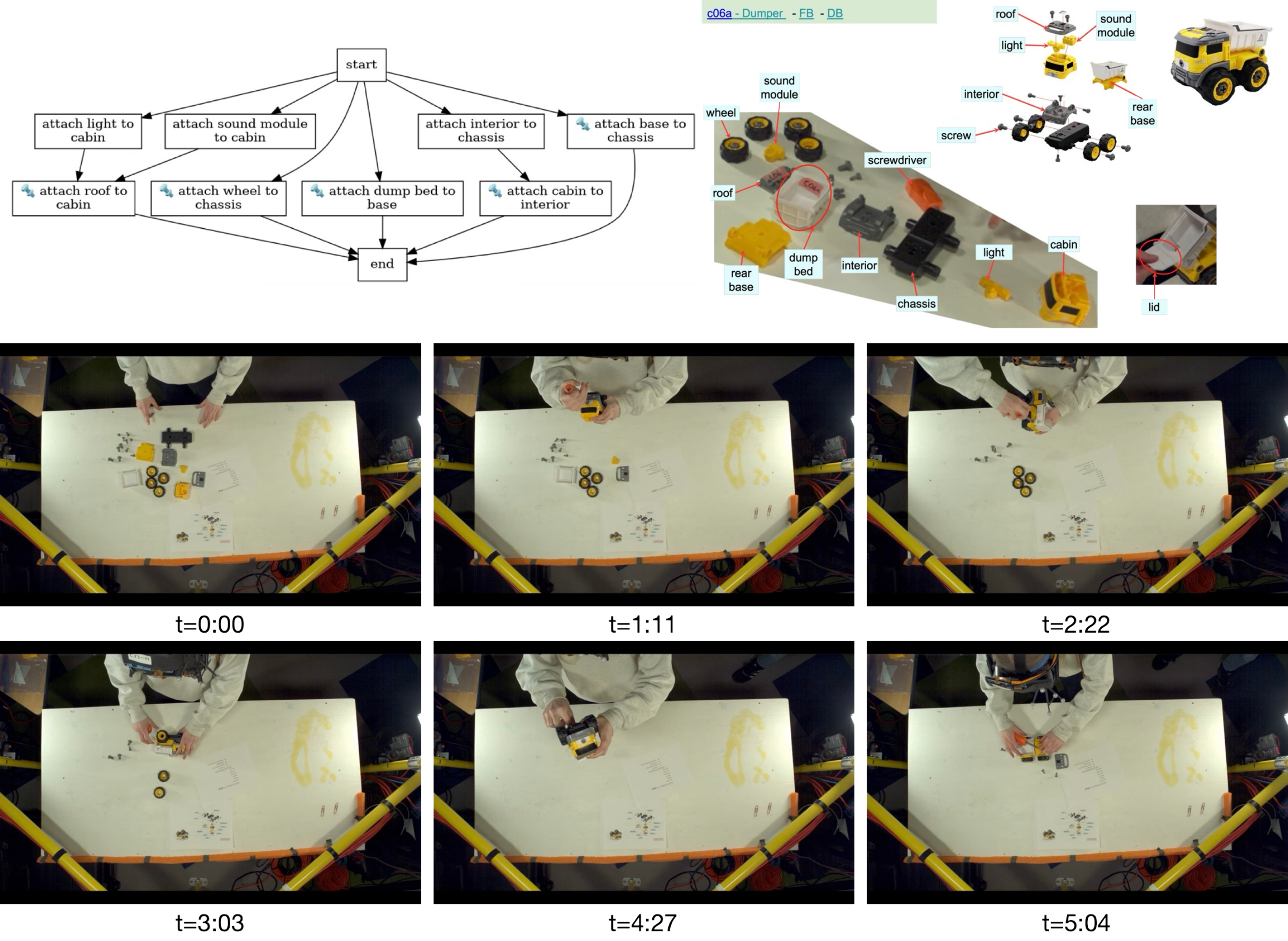}}\\
    Question: & What should I do next to fix my mistake? \\
    \multirow{4}{*}{Gold answer:} & You should detach the cabin from interior and secure the interior with screw and reattach the cabin. \\
    & Detach the roof, sound module, and cabin from the interior, so that you can screw in the interior. Then reattach everything in reverse order. \\
    \midrule
    DeepSeek-R1: & (... I need to figure out the correct next step based on the provided DOT graph. ... since the user's question is about fixing a mistake, the most logical next step would be to complete a prerequisite that was skipped. Looking at the graph, after starting, the user should attach the interior to the chassis, which then leads to attaching the cabin to the interior. So maybe they skipped attaching the interior first. ...) \\
    & \halmark~To fix your mistake, attach the interior to the chassis first. \\
    Qwen2.5-VL: & \xmark~You should attach the wheel to the chassis with a screw. \\
    GPT-4o: & \xmark~To fix your mistake, identify the last incorrect step in the assembly process by comparing your progress with the manual, disassemble the affected parts, and reassemble them in the correct order. \\
    Claude 3.7: & \xmark~Based on the recording, you need to attach the light to the cabin before attaching the roof, as you appear to have attached the roof without first installing the light component according to the assembly manual's dependency order. \\
    \bottomrule
\end{tabular}
\label{tab:qualitative-example2}
\end{table*}

\begin{figure*}
    \begin{lstlisting}[style=prompt]
# Instruction
You are playing with a take-apart toy alongside your senior friend, who has more experience with it.
You identify your skill level as intermediate.
Here is/are the step(s) you have already performed in the actual order:
* attach wheel to chassis
* attach interior to chassis (mistake: wrong position)
* detach interior from chassis
* attach interior to chassis (mistake: wrong position)

Your action history indicates some location errors.
While your friend is aware of these, you may or may not know.

# Task
What question would you ask your senior friend regarding the part location of the current (latest) step?

Create three possible QAs as a list. Questions should be something you would ask to your friend and the answers are something your friend would say.
Follow the format below:
* <question 1>
    * <answer 1>
    * ...
* ...

# Note
* Each question should require information about both A) what you have done and B) what you are supposed to do to answer correctly.
* Each question/answer should consist of one short and concise sentence/phrase.
* Your friend is watching over you, so you do not have to explain your situation, just ask questions directly.
* If multiple correct answers exist, provide all of them.
* Consider different personas you might adopt: careless/careful, child/teen/adult, casual/formal, etc.
* Return as diverse QA pairs as possible in terms of tone, wording, etc.
* Come up with different question types: yes/no-questions, Wh-questions, etc.

# Example
* Is X in the right place?
    * ...
* Where should I put this now?
    * ...
* ...

# Response
    \end{lstlisting}
    \caption{QA generation prompt example: \textit{``Default''} for ``location'' type.}
    \label{fig:prompt-full-qa-generation-default}
\end{figure*}
\begin{figure*}
    \begin{lstlisting}[style=prompt]
# Instruction                                                     
You are playing with a take-apart toy alongside your senior friend, who has more experience with it.                                                   
You identify your skill level as intermediate.                                               
Here is/are the step(s) you have already performed in the actual order (Note: the indented ones are the detailed, fine-grained steps):                 
* attach excavator arm to chassis (mistake: wrong position) (no-screw step)                  
    * position excavator arm                                                                 
* attach arm connector to excavator arm (mistake: previous one is a mistake) (screw-required step)                  
    * Pick up arm connector
    <<<omitted for brevity>>>
* detach excavator arm from chassis (no-screw step)
    * tilt up excavator arm
    <<<omitted for brevity>>>
* detach arm connector from excavator arm (no-screw step)
    * rotate excavator arm
The last step may be incomplete.

You do not have any missing steps either because you have followed the correct order or due to your previous mistake(s).
While your friend understand the correct orders, you may or may not know.

# Task
What question would you ask your senior friend regarding possible missing steps?
<<omitted for brevity>>

# Response
    \end{lstlisting}
    \caption{QA generation prompt example: \textit{``With fine''} for ``missing'' type. Note that some fine-grained actions are omitted for brevity.}
    \label{fig:prompt-full-qa-generation-with-fine}
\end{figure*}
\begin{figure*}
    \begin{lstlisting}[style=prompt]
# Instruction
You are playing with a take-apart toy alongside your senior friend, who has more experience with it.
You identify your skill level as intermediate.
Attached is the image of parts, final picture, and/or exploded view.
Here is/are the step(s) you have already performed in the actual order:
* attach interior to rear body (mistake: wrong position)
* attach cabin to interior (mistake: previous one is a mistake)

The above list contains your previous mistake information, if any.
While your friend is aware of these, you may or may not know.

# Task
What question would you ask your senior friend regarding the possible past mistakes?
<<omitted for brevity>>
# Response
    \end{lstlisting}
    \caption{
    QA generation prompt example: \textit{``With image''} for ``past'' type. 
    Note that a corresponding parts' image as shown in the lower middle of~\autoref{fig:overview} is also fed to a model.
    }
    \label{fig:prompt-full-qa-generation-with-image}
\end{figure*}
\begin{figure*}
    \begin{lstlisting}[style=prompt]
# Instruction
You are playing with a take-apart toy alongside your senior friend, who has more experience with it.
You identify your skill level as expert.
Here is/are the step(s) you have already performed in the actual order:
* attach wheel to chassis
* attach rear bumper to chassis
* attach cabin to interior

Below is the remaining part(s):
* bumper
* dump bed/base
* lid

# Task
What question would you ask your senior friend regarding next steps?

Create three possible questions as a list.
Follow the format below:
* <question 1>
* <question 2>
* <question 3>
<<omitted for brevity>>
# Response
    \end{lstlisting}
    \caption{
    QA generation prompt example: prompt for question generation in \textit{``Q then A''} for ``next'' type. 
    }
    \label{fig:prompt-full-qa-generation-q}
\end{figure*}
\begin{figure*}
    \begin{lstlisting}[style=prompt]
# Instruction
You are watching over your younger frined, who is playing with a take-apart toy. You are more experienced with it.
They identify their skill level as expert.
Your younger frined has done the following steps in this order:
* attach door to base
* attach transport cabin to base
<<omitted for brevity>>

Any of the steps below can be done as the possible next step:
* attach rear roof to transport cabin
<<omitted for brevity>>

# Task
Suppose you were asked the following question by your yonger friend:
Question: "Should I attach the bumper before the roof?"
What response(s) would you give them?

Create possible answers as a list, like:
* <answer 1>
* ...

# Note
* An answer should consist of one concise sentence/phrase.
* If multiple correct answers exist, provide all of them.

# Response
    \end{lstlisting}
    \caption{
    QA generation prompt example: prompt for answer generation in \textit{``Q then A''} for ``next'' type. 
    }
    \label{fig:prompt-full-qa-generation-a}
\end{figure*}

\begin{figure*}
    \begin{lstlisting}[style=prompt]
# Instruction                                                     
This is a multimodal question answering task.                                                
A user is assembling a toy car.                                                              
                                                                                             
# Parts/Final Picture/Exploded View                                                          
This is the image containing the parts, final picture, and/or exploded view.                 
<<<parts image added here>>>
                                                                            
# Assembling Manual                                                                          
This is the assembling manual as text in the DOT (graph description language from Graphviz) format.                                                                                       
Each node represents one step and each edge represents an order dependency.                  
Two nodes connected by an edge must be performed in the specified order.                     
Nodes that are not directly connected can be performed in any order, as long as their respective prerequisites have been completed.                                                       
digraph G {                                                                                  
start;
end;
"attach body to chassis w/ screw";
"attach interior to chassis w/ screw";
"attach roller to push frame w/ screw";
"attach roof to cabin w/ screw";
"attach wheel to chassis w/ screw";
"attach arm connector to chassis w/ screw";
"attach arm connector to push frame";
"attach bumper to body w/ screw";
"attach interior to cabin";
start -> "attach roller to push frame w/ screw";
start -> "attach interior to chassis w/ screw";
"attach wheel to chassis w/ screw" -> end;
"attach roller to push frame w/ screw" -> end; 
start -> "attach roof to cabin w/ screw";
"attach roof to cabin w/ screw" -> end;
start -> "attach wheel to chassis w/ screw";
start -> "attach arm connector to push frame"; 
"attach arm connector to push frame" -> "attach arm connector to chassis w/ screw";
"attach bumper to body w/ screw" -> end;
"attach arm connector to chassis w/ screw" -> end;
"attach body to chassis w/ screw" -> "attach bumper to body w/ screw";
"attach interior to cabin" -> "attach body to chassis w/ screw";
"attach interior to chassis w/ screw" -> "attach interior to cabin";
}

# Recording
These are the sampled frames in sequence from the recording of the user's activity.
<<<sampled frames added here>>>
# Task
The user asked the following question. Answer the question in one consice sentence, based on the give information above (parts, manual, and recording).
[Question]
Why is the rear body not fitting properly?
[Answer]
    \end{lstlisting}
    \caption{Benchmarking prompt example. Note that the parts image and sampled frames are omitted for brevity.}
    \label{fig:prompt-full-benchmark}
\end{figure*}
\begin{figure*}
    \begin{lstlisting}[style=prompt]
## Instruction ##                                                                                                                 
This is an evaluation task.                                                                                                                            
You will be given a question, gold answer(s), and predicted answer.                                                                                    
Your task is to evaluate if the predicted answer matches against the gold answer(s).                                                                   

Here is/are the step(s) they have already performed in the actual order:                                                                               
* attach wheel to chassis                                                                                                                              
* attach base to chassis                                                                                                                               
* detach base from chassis                                                                                                                             
* attach interior to chassis                                                                                                                           

Give your ternary judge 0, 1, or 2:                                                                                                                   
* 0 means the predicted answer is wrong (unmatch)                                                                                                      
* 1 means the predicted answer is partially correct/wrong (partial match)                                                                              
* 2 means the predicted answer is correct (match)                                                                                                     
When multiple gold answers are available (provided as a list), the predicted answer is correct/partially correct if it matches/partially matches with at least one of the gold answers.                                                                                                                     

Provide your feedback as follows:                                                                                                                      
## Feedback ##                                                                                                                                         
[Rationale] (your rationale for the judge, as a text)                                                                                                  
[Judge] (your judge, as a number, 0, 1, or 2)                                                                                                         

## Note ##                                                                                                                                            
The question is being asked by a user who is playing with a take-apart toy.                                                                            
Gold answer(s) are created by well-trained humans.                                                                                                     
Predicted answer is created by a machine, based on the corresponding instruction and the frames of the assemblying process recording.

## Task ##                                                                                                                                             
Now, here are the question, gold answer(s), and predicted answer:                                                                                      
[Question]                                                                                                                                             
Am I missing a step before attaching the interior?                                                                                                     
[Gold Answer(s)]                                                                                                                                       
- Yes, you need to secure the rear base.
- No, you're on the right track.
[Predicted Answer]                                                                                                                                     
Yes, you need to attach the sound module to the chassis before attaching the interior.                                                                 

## Feedback ##                                                                                                                                         
[Rationale]                                                                                                                                            
    \end{lstlisting}
    \caption{LLM-as-a-judge prompt example.}
    \label{fig:prompt-full-benchmark-eval}
\end{figure*}

\end{document}